\documentclass{article}

\usepackage{times}
\usepackage{graphicx}
\usepackage{subfig}
\usepackage{natbib}
\usepackage{amsmath,amssymb}
\usepackage[bold]{hhtensor}

\usepackage{hyperref}

\graphicspath{{./figure/}}

\usepackage[accepted]{icml2016}

\icmltitlerunning{Minimal Gated Unit for Recurrent Neural Networks}

\begin{document} 

\twocolumn[
\icmltitle{Minimal Gated Unit for Recurrent Neural Networks}

\icmlauthor{Guo-Bing Zhou}{zhougb@lamda.nju.edu.cn}
\icmlauthor{Jianxin Wu}{wujx@nju.edu.cn}
\icmlauthor{Chen-Lin Zhang}{u-zhangcl@lamda.nju.edu.cn}
\icmlauthor{Zhi-Hua Zhou}{zhouzh@nju.edu.cn}
\icmladdress{National Key Laboratory for Novel Software Technology,
               Nanjing University, Nanjing, China, 210023}

\icmlkeywords{Recurrent neural networks, LSTM, GRU, minimal gated unit}

\vskip 0.3in
]

\begin{abstract}
Recently recurrent neural networks (RNN) has been very successful in handling sequence data. However, understanding RNN and finding the best practices for RNN is a difficult task, partly because there are many competing and complex hidden units (such as LSTM and GRU). We propose a gated unit for RNN, named as Minimal Gated Unit (MGU), since it only contains one gate, which is a minimal design among all gated hidden units. The design of MGU benefits from evaluation results on LSTM and GRU in the literature. Experiments on various sequence data show that MGU has comparable accuracy with GRU, but has a simpler structure, fewer parameters, and faster training. Hence, MGU is suitable in RNN's applications. Its simple architecture also means that it is easier to evaluate and tune, and in principle it is easier to study MGU's properties theoretically and empirically.
\end{abstract}

\section{Introduction}\label{sec:intro}

In recent years, deep learning models have been particularly effective in dealing with data that have complex internal structures. For example, convolutional neural networks (CNN) are very effective in handling image data in which 2D spatial relationships are critical among the set of raw pixel values in an image~\cite{lr:LeCun1998,vi:Krizhevsky2012}. Another success of deep learning is handling sequence data, in which the sequential relationships within a variable length input sequence is crucial. In sequence modeling, recurrent neural networks (RNN) have been very successful in language translation~\cite{lr:Cho2014_GRU,lr:Sutskever2014_Translation,lr:Bahdanau2015_ICLR}, speech recognition~\cite{lr:Graves2013_Speech}, image captioning, i.e., summarizing the semantic meaning of an image into a sentence~\cite{vi:Xu2015_captioning,vi:Karparthy2015_Captioning,lr:Lebret2015_Captioning}, recognizing actions in videos~\cite{vi:Donahue2015_Video,lr:Srivastava2015_Video_LSTM}, or short-term precipitation prediction~\cite{lr:Shi2015_ConvLSTM}.

There is, however, one important difference between RNN and CNN. The key building blocks of CNN, such as nonlinear activation function, convolution and pooling operations, etc., have been extensively studied. The choices are becoming \emph{convergent}, e.g., ReLU for nonlinear activation, small convolution kernels and max-pooling. Visualization also help us understand the semantic functionalities of different layers~\cite{vi:Zeiler2014}, e.g., firing at edges, corners, combination of specific edge groups, object parts, and objects.

The community's understanding of RNN, however, is not as thorough, and the opinions are much less convergent. Proposed by Hochreiter and Schmidhuber~\yrcite{lr:Hochreiter1997_LSTM}, the Long Short-Term Memory (LSTM) model and its variants have been the overall best performing RNN. Typical LSTM is complex, having 3 gates and 2 hidden states. Using LSTM as an representative example, we may find some obstacles that prevent us from reaching a consensus on RNN:
\begin{itemize}
 \item \textbf{Depth with unrolling.} Different from CNN which has physical layers, RNN is considered a deep model only when it is unrolled along the time axis. This property helps in reducing parameter size, but at the same time hinders visualization and understanding. This difficulty comes from the complexity of sequence data itself. We do not have a satisfactory solution for solving this difficulty, but we hope this paper will help in mitigating difficulties caused by the next one.
 \item \textbf{Competing and complex structures.} Since its inception, the focus of LSTM-related research has been altering its structure to achieve higher performance. For example, a forget gate was added by~\citet{lr:Gers1999_ForgetGate} to the original LSTM, and a peephole connection made it even more complex~\cite{lr:Gers2002_PeepholeLSTM}. It was not until recently that simplified models were proposed and studied, such as the Gated Recurrent Unit (GRU) by~\citet{lr:Cho2014_GRU}. GRU, however, is still relatively complex because it has two gates. Very recently there have been empirical evaluations on LSTM, GRU, and their variants~\cite{lr:Chung2014_Eval_GRU,lr:Jozefowicz2015_Eval_LSTM,lr:Greff2015_Search_LSTM}. Unfortunately, no consensus has yet been reached on the best LSTM-like RNN architecture.
\end{itemize}

Theoretical analysis and empirical understanding of deep learning techniques are fundamental. However, it is very difficult if there are too many components in the structure. Simplifying the model structure is an important step to enable the learning theory analysis in the future.

Complex RNN models not only hinder our understanding. It also means that more parameters are to be learned and more components to be tuned. As a natural consequence, more training sequences and (perhaps) more training time are needed. However, evaluations in~\cite{lr:Chung2014_Eval_GRU} and~\cite{lr:Jozefowicz2015_Eval_LSTM} both show that more gates do not lead to better accuracy. On the contrary, the accuracy of GRU is usually higher than that of LSTM, albeit the fact that GRU has one less hidden state and one less gate than LSTM.

In this paper, we propose a new variant of GRU (which is also a variant of LSTM), which has minimal number of gates--only one gate! Hence, the proposed method is named as the Minimal Gated Unit (MGU). Evaluations in~\cite{lr:Chung2014_Eval_GRU,lr:Jozefowicz2015_Eval_LSTM,lr:Greff2015_Search_LSTM} agreed that RNN with a gated unit works significantly better than a RNN with a simple $\tanh$ unit without any gate. The proposed method has the smallest possible number of gates in any gated unit, a fact giving rise to the name \emph{minimal} gated unit.

With only one gate, we expect MGU will have significantly fewer parameters to learn than GRU or LSTM, and also fewer components or variations to tune. The learning process will be faster compared to them, which will be verified by our experiments on a diverse set of sequence data in Section~\ref{sec:results}. What is more, our experiments also showed that MGU has overall comparable accuracy with GRU, which once more concurs the observation that fewer gates reduces complexity but not necessarily accuracy.

Before we present the details of MGU, we want to add that we are not proposing a ``better'' RNN model in this paper.\footnote{We also believe that: without a carefully designed common set of comprehensive benchmark datasets and evaluation criteria, it is not easy to get conclusive decisions as to which RNN model is better.} The purpose of MGU is two-fold. First, with a simpler model, we can reduce the requirement for training data, architecture tuning and CPU time, while at the same time maintaining accuracy. This characteristic may make MGU a good candidate in various applications. Second, a minimally designed gated unit will (in principle) make our analyses of RNN easier, and help us understand RNN, but this will be left as a future work.

\section{RNN: LSTM, GRU, and More}

We start by introducing various RNN models, mainly LSTM and GRU, and their evaluation results. These studies in the literature have guided us in how to minimize a gated hidden unit in RNN.

A recurrent neural network uses an index $t$ to indicate different positions in an input sequence, and assumes that there is a hidden state $\vec{h}_t$ to represent the system status at time $t$.\footnote{We use boldface letters to denote vectors.} RNN accepts input $\vec{x}_t$ at time $t$, and the status is updated by a nonlinear mapping $f$ from time to time: 
\begin{equation}
 \vec{h}_{t} = f(\vec{h}_{t-1},\vec{x}_t) \,.
\end{equation}
One usual way of defining the recurrent unit $f$ is a linear transformation plus a nonlinear activation, e.g.,
\begin{equation}
 \vec{h}_t = \tanh\left( W \left[ \vec{h}_{t-1},\vec{x}_t\right] + \vec{b} \right)  \,,
\end{equation}
where we combined the parameters related to $\vec{h}_{t-1}$ and $\vec{x}_t$ into a matrix $W$, and $\vec{b}$ is a bias term. The activation ($\tanh$) is applied to every element of its input. The task of RNN is to learn the parameters $W$ and $\vec{b}$, and  we call this architecture the simple RNN. An RNN may also have an optional output vector $\vec{y}_t$.

\textbf{LSTM.} RNN in the simple form suffers from the vanishing or exploding gradient issue, which makes learning RNN using gradient descent very difficult in long sequences~\cite{lr:Bengio1994_RNN_Gradient,lr:Hochreiter1997_LSTM}. LSTM solved the gradient issue by introducing various gates to control how information flows in RNN, which are summarized in Table~\ref{tbl:equations}, from Equation~(\ref{eqn:LSTM_forget}) to (\ref{eqn:LSTM_hidden}), in which 
\begin{equation}
 \sigma(x)=\frac{1}{1+\exp(-x)}
\end{equation}
is the logistic sigmoid function (applied to every component of the vector input) and $\odot$ is the component-wise product between two vectors.

\begin{table}
 \caption{Summary of three gated units: LSTM, GRU, and the proposed MGU. $\sigma$ is the logistic sigmoid function, and $\odot$ means component-wise product.}\label{tbl:equations}
 \begin{tabular}{|l|}
  \hline
  LSTM (Long Short-Term Memory)
  \\ \hline
  \parbox[b]{0.95\columnwidth}{
      \vspace{-8pt}
      \begin{subequations}
	\begin{align}
	  &\vec{f}_t = \sigma\left(W_f \left[ \vec{h}_{t-1}, \vec{x}_t \right]+\vec{b}_f \right) \,,\qquad\quad&\label{eqn:LSTM_forget} \\
	  &\vec{i}_t = \sigma\left(W_i \left[ \vec{h}_{t-1}, \vec{x}_t \right]+\vec{b}_i \right) \,, \label{eqn:LSTM_input}\\
	  &\vec{o}_t = \sigma\left(W_o \left[ \vec{h}_{t-1}, \vec{x}_t \right]+\vec{b}_o \right) \,, \label{eqn:LSTM_output}\\
	  &\tilde{\vec{c}}_t = \tanh\left(W_c \left[ \vec{h}_{t-1}, \vec{x}_t \right]+\vec{b}_c \right) \,, \label{eqn:LSTM_temp_cell}\\
	  &\vec{c}_t = \vec{f}_t \odot \vec{c}_{t-1} + \vec{i}_t \odot \tilde{\vec{c}}_t \,, \label{eqn:LSTM_cell}\\
	  &\vec{h}_t = \vec{o}_t \odot \tanh(\vec{c}_t) \,.\label{eqn:LSTM_hidden}
	\end{align}
      \end{subequations} 
      \vspace{-16pt}} 
  \\ \hline \hline
  GRU (Gated Recurrent Unit)
  \\ \hline
  \parbox[b]{0.95\columnwidth}{
      \vspace{-8pt}
      \begin{subequations}
	\begin{align}
	  &\vec{z}_t = \sigma\left(W_z \left[ \vec{h}_{t-1}, \vec{x}_t \right]+\vec{b}_z \right) \,,\qquad\quad&\label{eqn:GRU_update} \\
	  &\vec{r}_t = \sigma\left(W_r \left[ \vec{h}_{t-1}, \vec{x}_t \right]+\vec{b}_r \right) \,,\label{eqn:GRU_reset} \\
	  &\tilde{\vec{h}}_t = \tanh\left(W_h \left[ \vec{r}_t \odot \vec{h}_{t-1}, \vec{x}_t \right]+\vec{b}_h \right) \,, \label{eqn:GRU_temp_hidden} \\
	  &\vec{h}_t = (1-\vec{z}_t) \odot \vec{h}_{t-1} + \vec{z}_t \odot \tilde{\vec{h}}_t \,. \label{eqn:GRU_hidden}
	\end{align}
      \end{subequations} 
      \vspace{-16pt}}
  \\ \hline \hline
  MGU (Minimal Gated Unit, the proposed method)
  \\ \hline
  \parbox[b]{0.95\columnwidth}{
      \vspace{-8pt}
      \begin{subequations}
	\begin{align}
	  &\vec{f}_t = \sigma\left(W_f \left[ \vec{h}_{t-1}, \vec{x}_t \right]+\vec{b}_f \right) \,,\qquad\quad&\label{eqn:MGU_forget} \\
	  &\tilde{\vec{h}}_t = \tanh\left(W_h \left[ \vec{f}_t \odot \vec{h}_{t-1}, \vec{x}_t \right]+\vec{b}_h \right) \,, \label{eqn:MGU_temp_hidden} \\
	  &\vec{h}_t = (1-\vec{f}_t) \odot \vec{h}_{t-1} + \vec{f}_t \odot \tilde{\vec{h}}_t \,. \label{eqn:MGU_hidden}
	\end{align}
      \end{subequations} 
      \vspace{-16pt}}
  \\ \hline
 \end{tabular}
\end{table}

\begin{figure}
 \centering
 \subfloat[Long Short-Term Memory (LSTM)]{\includegraphics[width=\columnwidth]{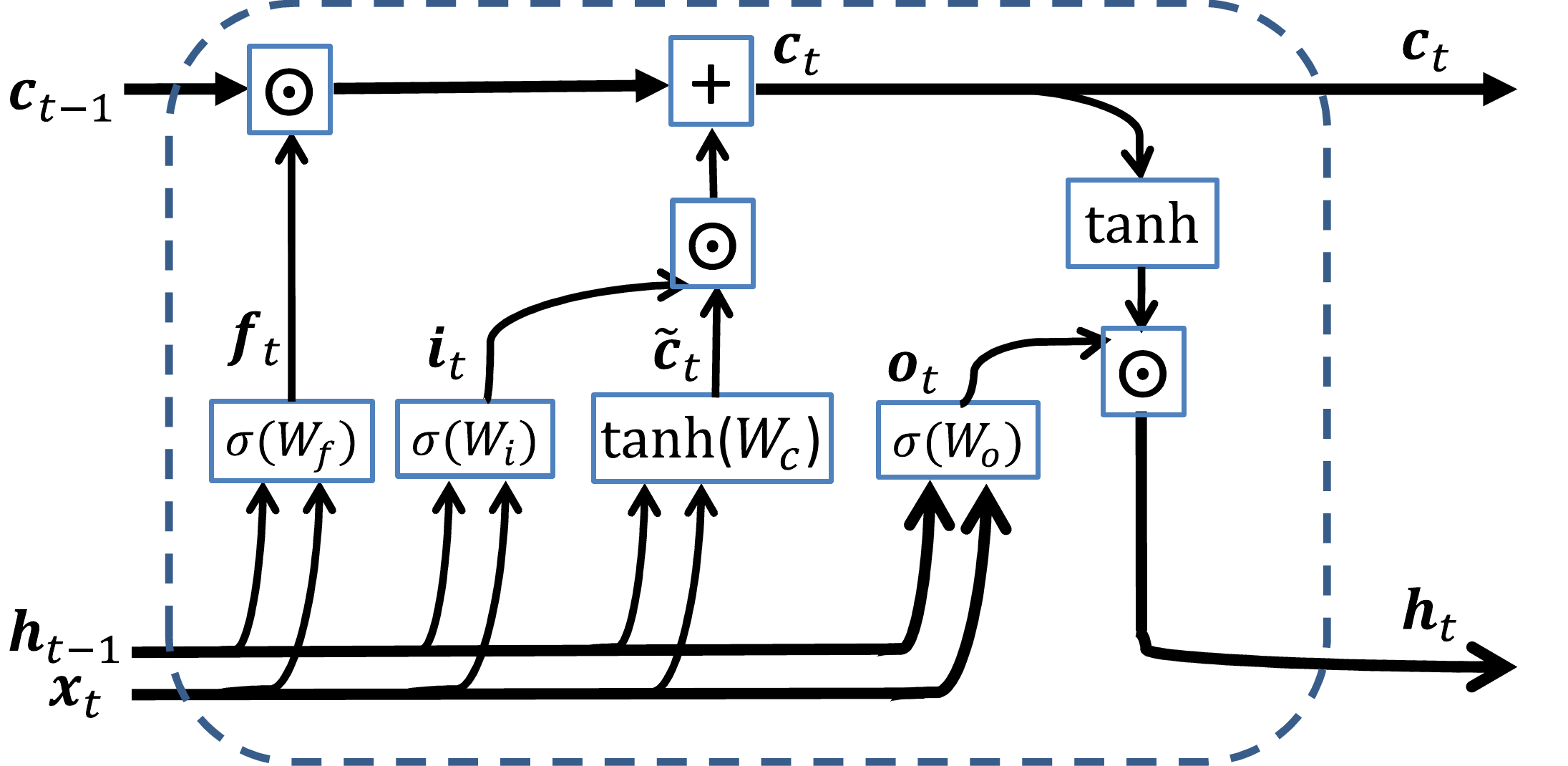}\label{fig:LSTM}} \hspace{12pt}
 \subfloat[Coupled LSTM]{\includegraphics[width=\columnwidth]{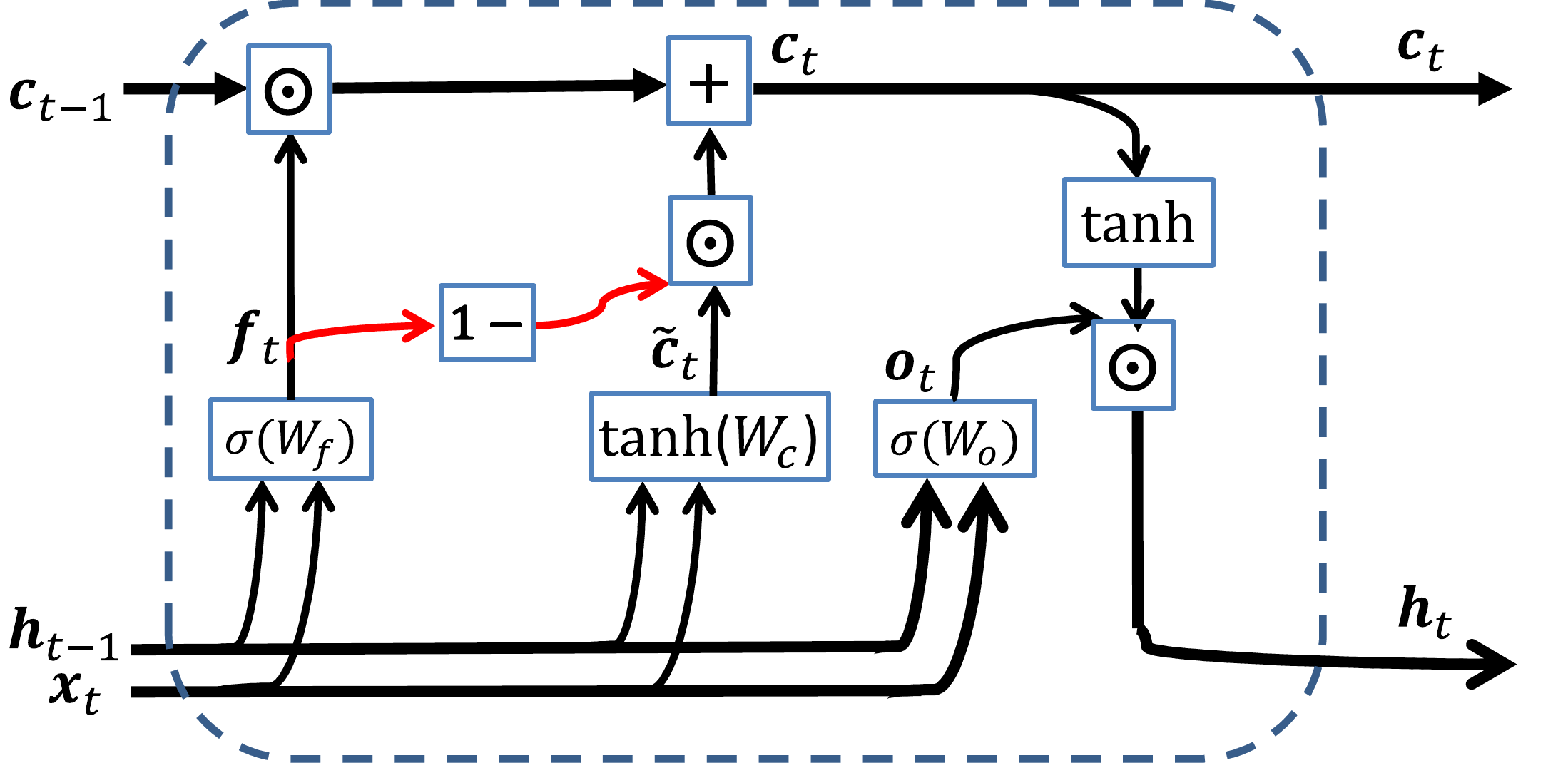}\label{fig:coupled_LSTM}} \\
 \subfloat[Gated Recurrent Unit (GRU)]{\includegraphics[width=\columnwidth]{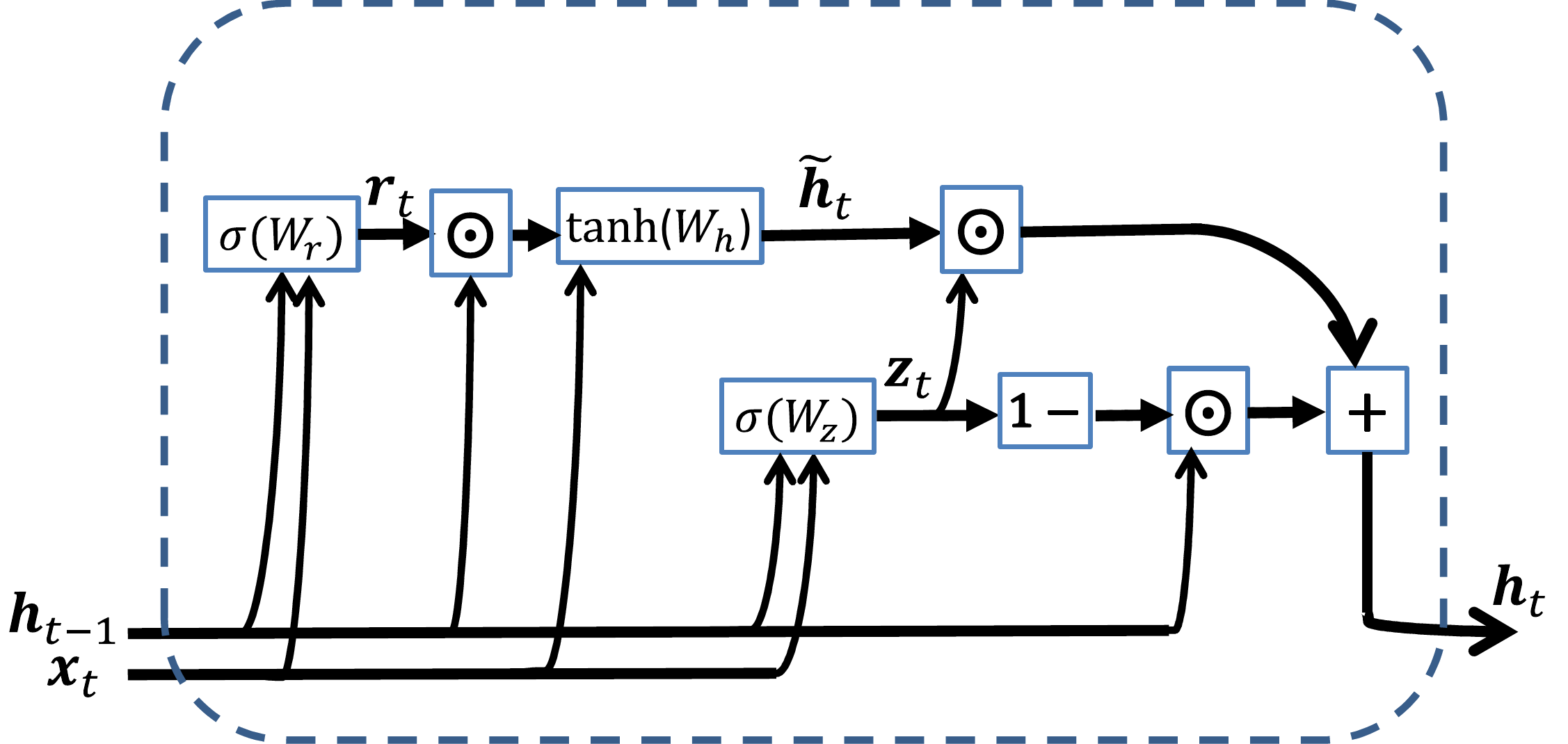}\label{fig:GRU}} \hspace{12pt}
 \subfloat[Minimal Gated Unit (MGU, the proposed method)]{\includegraphics[width=\columnwidth]{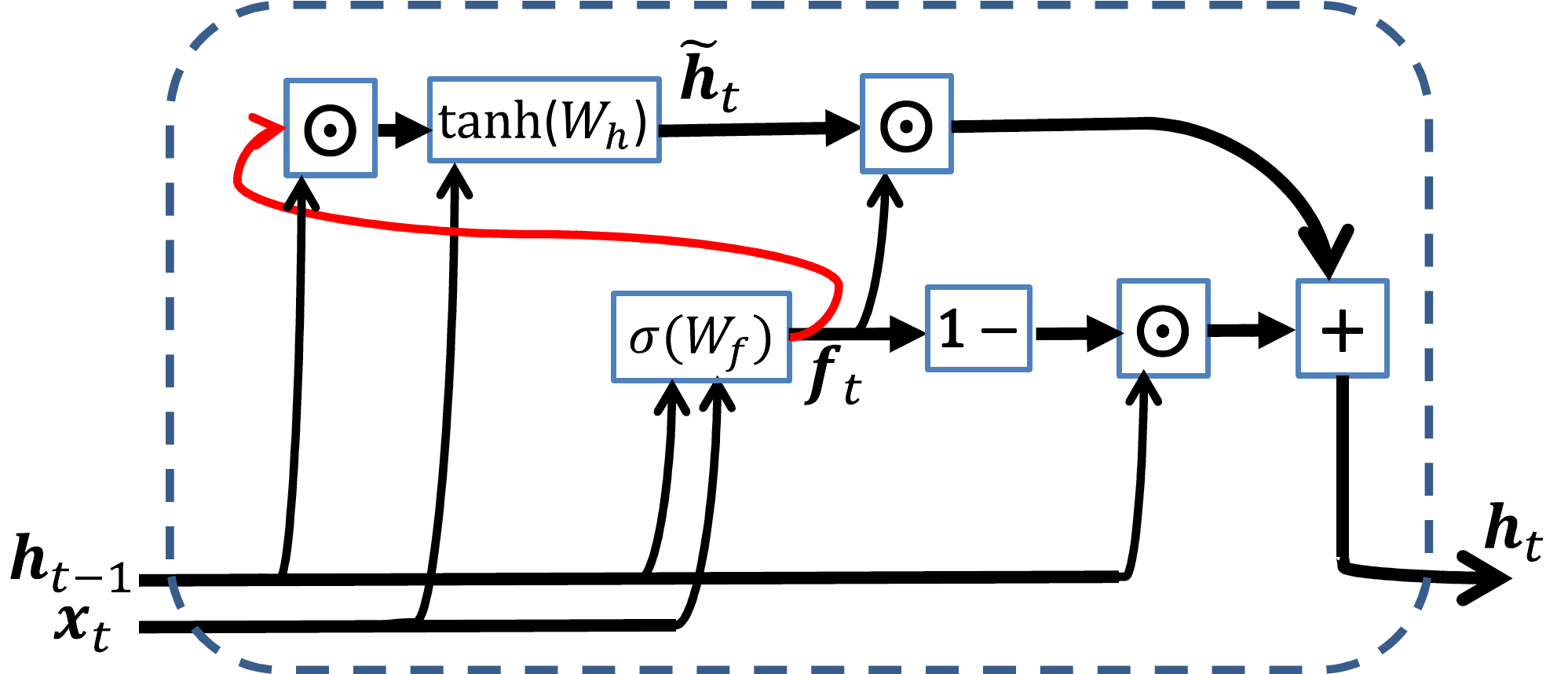}\label{fig:MGU}}
 \caption{Data flow and operations in various gated RNN models. The direction of data flow are indicated by arrows, and operations on data are shown in rectangles. Five types of element wise operations (logistic sigmoid, $\tanh$, plus, product and one minus) are involved. For operations with parameters (logistic sigmoid and $\tanh$), we also include their parameters in the rectangle. These figures are different from diagrams that illustrate gates as switches, but match better to the equations in Table~\ref{tbl:equations}.} \label{fig:flow}
\end{figure}

Figure~\ref{fig:LSTM} illustrates the data flow and operations in LSTM.
\begin{itemize}
 \item There is one more hidden state $\vec{c}_t$ in addition to $\vec{h}_t$, which helps maintain long-term memories.
 \item The forget gate $\vec{f}_t$ decides the portion (between 0 and 1) of $\vec{c}_{t-1}$ to be remembered, determined by parameters $W_f$ and $\vec{b}_f$.
 \item The input gate $\vec{i}_t$ determines which portion of time $t$'s new information is to be added to $\vec{c}_t$ with parameter $W_i$ and $\vec{b}_i$.
 \item The inputs are transformed as the update $\tilde{\vec{c}}_t$ with parameters $W_c$ and $\vec{b}_c$. $\tilde{\vec{c}}_t$ (weighted by $\vec{i}_t$) and $\vec{c}_{t-1}$ (weighted by $\vec{f}_t$)  form the new cell state $\vec{c}_t$.
 \item An output gate $\vec{o}_t$ is determined by parameters $W_o$ and $\vec{b}_o$, and controls which part of $\vec{c}_t$ is to be output as the hidden state $\vec{h}_t$.
\end{itemize}
LSTM learning follows typical stochastic gradient descent and back propagation~\cite{lr:Graves2005_FullGraident}.

There are a lot of variants of LSTM. The original LSTM \cite{lr:Hochreiter1997_LSTM} does not include the forget gate, which was later introduced in~\cite{lr:Gers1999_ForgetGate}. \citet{lr:Gers2002_PeepholeLSTM} makes LSTM even more complicated by allowing the three gates ($\vec{f}_t$, $\vec{i}_t$, $\vec{o}_t$) to take $\vec{c}_{t-1}$ or $\vec{c}_t$ as an additional input, called the peephole connections. We choose the specific form of LSTM in Table~\ref{tbl:equations} because of two reasons. First, as will soon be discussed, the forget gate is essential. Second, the peephole connection does not seem necessary, but it complicates the learning process.

However, recently the trend is reversed: researchers found that simplifying the LSTM architecture may improve its performance.

\textbf{LSTM with coupled gates.} \citet{lr:Greff2015_Search_LSTM} evaluated a variant of LSTM, which couples the forget and input gates into one:
\begin{equation}
 \vec{i}_t = 1 - \vec{f}_t\,, \;\; \forall t \,,
\end{equation}
as illustrated in Figure~\ref{fig:coupled_LSTM}. The coupling removed one gate and its parameters ($W_i$ and $\vec{b}_i$), which leads to reduced computational complexity and slightly higher accuracy. \citet{lr:Greff2015_Search_LSTM} also observed that removing the peephole connection has similar effects.

\textbf{GRU.} The Gated Recurrent Unit (GRU) architecture further simplifies LSTM-like units~\cite{lr:Cho2014_GRU}. GRU contains two gates: an update gate $\vec{z}$ (whose role is similar to the forget gate) and a reset gate $\vec{r}$ (whose role loosely matches the input gate). GRU's update rules are shown as Equation~(\ref{eqn:GRU_update}) to (\ref{eqn:GRU_hidden}), and the data flow and operations are illustrated in Figure~\ref{fig:GRU}. Beyond removing the output gate from LSTM, GRU also removed the hidden (slowly-changing) cell state $\vec{c}$. Note that GRU has appeared in different forms. When~\citet{lr:Cho2014_GRU} originally proposed GRU, its form is different from Equation~\ref{eqn:GRU_hidden}, as
\begin{equation}
 \vec{h}_t = \vec{z}_t \odot \vec{h}_{t-1} + (1-\vec{z}_t) \odot \tilde{\vec{h}}_t \,. 
\end{equation}
However, these two forms are mathematically equivalent. We adopt Equation~\ref{eqn:GRU_hidden} because it is more popular in the literature. 

Evaluations in~\cite{lr:Chung2014_Eval_GRU} found that when LSTM and GRU have the same amount of parameters, GRU slightly outperforms LSTM. Similar observations were also corroborated in~\cite{lr:Jozefowicz2015_Eval_LSTM}.

\textbf{SCRN.} Instead of using gates to control information flow in RNN, the Structurally Constrained Recurrent Network (SCRN) added a hidden context vector $\vec{s}_t$ to simple RNN, which changes slowly over time if the parameter $\alpha$ is large~\cite{lr:Mikolov2015_SCRN}, as
\begin{align}
 \vec{s}_t &= \alpha \vec{s}_{t-1} + (1-\alpha)B\vec{x}_t \,, \\
 \vec{h}_t &= \sigma(P\vec{s}_t+A\vec{x}_t+R\vec{h}_{t-1}) \,,
\end{align}
in which $\alpha$ was fixed to 0.95, and the hidden state $\vec{h}_t$ hinges on three factors: $\vec{s}_t$, $\vec{x}_t$ and $\vec{h}_{t-1}$. SCRN has still fewer parameters than GRU, and has shown similar performance as LSTM in~\cite{lr:Mikolov2015_SCRN}.

\textbf{IRNN.} \citet{lr:Le2015_IRNN} showed that minor changes to the simple RNN architecture can significantly improve its accuracy. The key is to initialize the simple RNN's weight matrix to an identity matrix, and use ReLU (rectified linear unit) as the nonlinear activation function. This method (named as IRNN) has achieved accuracy that is much closer to LSTM than that of simple RNN. Especially in the MNIST dataset~\cite{lr:LeCun1998}, IRNN significantly outperforms LSTM. Similarly, \citet{lr:Jozefowicz2015_Eval_LSTM} also showed that proper initialization is also important for LSTM. They showed that the bias of the forget gate should be set to a large value (e.g., 1 or 2). The same initialization trick was used in~\cite{lr:Le2015_IRNN} too.

\textbf{LSTM variants.} \citet{lr:Greff2015_Search_LSTM} proposed, in evaluation of the importance of LSTM components, 8 variants of the LSTM architecture. The evaluation results, however, showed that none of them can outperform the vanilla LSTM model. Their vanilla LSTM architecture has the peephole connections. Hence, it is slightly different from the LSTM architecture used in this paper (cf. Table~\ref{tbl:equations}).

\textbf{GRU variants.} \citet{lr:Jozefowicz2015_Eval_LSTM} proposed three variants of GRU. In these variants, they add the $\tanh$ nonlinearity in generating the gates, removing dependency to the hidden state in generating the gates, and make these changes while generating $\tilde{\vec{h}}_t$, etc. These variants sometimes achieve higher accuracy than GRU, but none of them can consistently outperform GRU.

Overall, with all these recent results and proposals we feel that among them GRU has some advantage in learning recurrent neural networks. Although it is not always the model with the highest accuracy or fewest parameters, it has stable performance (and it is usually one of the most accurate models) and relatively small amount of parameters. We will use GRU as the baseline method and compare it with the proposed MGU (minimal gated unit, cf. next section) in our experiments.

\section{Minimal Gated Unit}

As introduced in Section~\ref{sec:intro}, we prefer an RNN architecture that has the smallest number of gates without losing LSTM's accuracy benefits. However, the choice of which gate to keep is no easy job. Fortunately, several recent evaluations have helped us make this decision. Now we briefly summarize knowledge from these evaluations.
\begin{description}
 \item[\citet{lr:Jozefowicz2015_Eval_LSTM}:] the forget gate is critical (and its biases $\vec{b}_f$ must be initialized to large values); the input gate is important, but the output gate is unimportant; GRU and LSTM have similar performance.
 \item[\citet{lr:Greff2015_Search_LSTM}:] The forget and output gates are critical, and many variants of LSTM (mainly simplified LSTM variants) act similarly to LSTM.
 \item[\citet{lr:Chung2014_Eval_GRU}:] Gated units work better than simple units without any gate; GRU and LSTM has comparable accuracy with the same number of parameters.
\end{description}

One notable thing is that different evaluations may lead to inconsistent conclusions, e.g., on the importance of the output gate~\citet{lr:Jozefowicz2015_Eval_LSTM} and~\citet{lr:Greff2015_Search_LSTM} disagreed. This is inevitable because data with different properties have been used in different evaluations. However, at least we find the following consensus among these evaluations: 
\begin{itemize}
 \item Having a gated unit is key to high performance of RNN architectures;
 \item The forget gate is unanimously considered the most important one; and, 
 \item A simplified model may lower complexity and maintain comparable accuracy.
\end{itemize}

Hence, we propose the Minimal Gated Unit (MGU), which has the smallest possible number of gates in any gated unit. MGU only has 1 gate, which is the forget gate. MGU is based on GRU, and it further couples the input (reset) gate to the forget (update) gate, by specifying that
\begin{equation}
 \vec{r}_t = \vec{f}_t \,, \;\; \forall t \,.
\end{equation}
Note that we use $\vec{f}$ (instead of $\vec{z}$) to denote the only gate, because it is treated as the forget gate (which can be considered as a renaming of the update gate $\vec{z}$ in GRU). The equations that define MGU are listed in Table~\ref{tbl:equations} as Equations~(\ref{eqn:MGU_forget}) to (\ref{eqn:MGU_hidden}), and the data flow and operations are illustrated in Figure~\ref{fig:MGU}.

In MGU, the forget gate $\vec{f}_t$ is first generated, and the element-wise product between $1-\vec{f}_t$ and $\vec{h}_{t-1}$ becomes part of the new hidden state $\vec{h}_t$. The portion of $\vec{h}_{t-1}$ that is ``forgotten'' ($\vec{f}_t \odot \vec{h}_{t-1}$) is combined with $\vec{x}_t$ to produce $\tilde{\vec{h}}_t$, the short-term response. A portion of $\tilde{\vec{h}}_t$ (determined again by $\vec{f}_t$) form the second part of $\vec{h}_t$.

Comparing the equation sets in Table~\ref{tbl:equations} and the parameterized operations in Figure~\ref{fig:flow}, it is obvious that MGU is more simplified than LSTM or GRU. While LSTM has four sets of parameters that determine $\vec{f}$, $\vec{i}$, $\vec{o}$ and $\tilde{\vec{c}}$, and GRU has three sets of parameters for $\vec{z}$, $\vec{r}$ and $\tilde{\vec{h}}$, MGU only has two sets of parameters, one for calculating $\vec{f}$, the other for $\tilde{\vec{h}}$. In other words, MGU has only roughly half the parameter size of that of LSTM, and 67\% of that of GRU, because $W_f$ (or $W_z$), $W_i$ (or $W_r$), $W_o$, $W_h$ have the same size. MGU also has slightly more parameters than SCRN, as will be shown in the example in Section~\ref{sec:results_PTB}. Since the parameter size of IRNN is the same as that of simple RNN, it must have the smallest number of parameters.

In short, MGU is a minimal design in any gated hidden unit for RNN. As we will show in the next section by experiments on a variety of sequence data, MGU also learns RNN for sequences without suffering from the gradient vanishing or gradient exploding problem (thanks to the forget gate in MGU).  Because MGU only has few factors to tune, it is easier to find the best practices for MGU than for other gated units.

\section{Experimental Results} \label{sec:results}

In this section, we will evaluate the effectiveness of MGU using four datasets. The simple adding problem is used as a sanity check in Section~\ref{sec:results_adding}. The IMDB dataset and the MNIST dataset are sentiment and image classification problems with sequence inputs, presented in Section~\ref{sec:results_imdb} and~\ref{sec:mnist}, respectively. Finally, we evaluate MGU on the Penn TreeBank (PTB) language modeling dataset in Section~\ref{sec:results_PTB}.

As was shown in the evaluations~\cite{lr:Chung2014_Eval_GRU,lr:Jozefowicz2015_Eval_LSTM}, GRU has comparable accuracy with LSTM, and has fewer parameters than LSTM. We will use GRU as a baseline architecture, and compare the proposed MGU with GRU. If not otherwise specified, we compare these two algorithms with the same number of hidden units. All RNNs are implemented with the Lasagne package in the Theano library.\footnote{\href{http://deeplearning.net/software/theano/}{http://deeplearning.net/software/theano/},\\ \phantom{abc\,} \href{http://lasagne.readthedocs.org}{http://lasagne.readthedocs.org}.}

The dropout technique is not used in either GRU or MGU. Because the focus of this paper is not absolutely high accuracy, we did not evaluate model averaging (ensemble) either.

The metrics that are compared include accuracy (or error, or perplexity) computed from the test sets, the average running time per epoch, and the number of parameters in the hidden units. We only count the parameters of the hidden unit, i.e., the preprocessing and fully connected regression or classification layer's parameters are not counted.

\subsection{The Adding Problem} \label{sec:results_adding}

The adding problem was originally proposed in~\cite{lr:Hochreiter1997_LSTM}, and we use the variant in~\cite{lr:Le2015_IRNN}. The input has two components: one random number in the range $[0\;1]$, and the other is a mask in $\{+1,0,-1\}$. In the sequence (whose length ranges from 50 to 55), only 2 numbers are with mask $+1$, and the output should be the sum of these two. Both MGU and GRU use 100 hidden units; batch size is 100, and the learning rate is $10^{-3}$. For this problem, we use a bidirectional network structure~\cite{lr:Graves2005_FullGraident}, which scans the sequence both from left to right and from right to left; the overall hidden state is the concatenation of the hidden state in both scans. On top of the last time step's hidden state, we add a fully connected layer to transform the last hidden state vector to a regression prediction.

\begin{figure}
 \centering
 \includegraphics[width=\columnwidth]{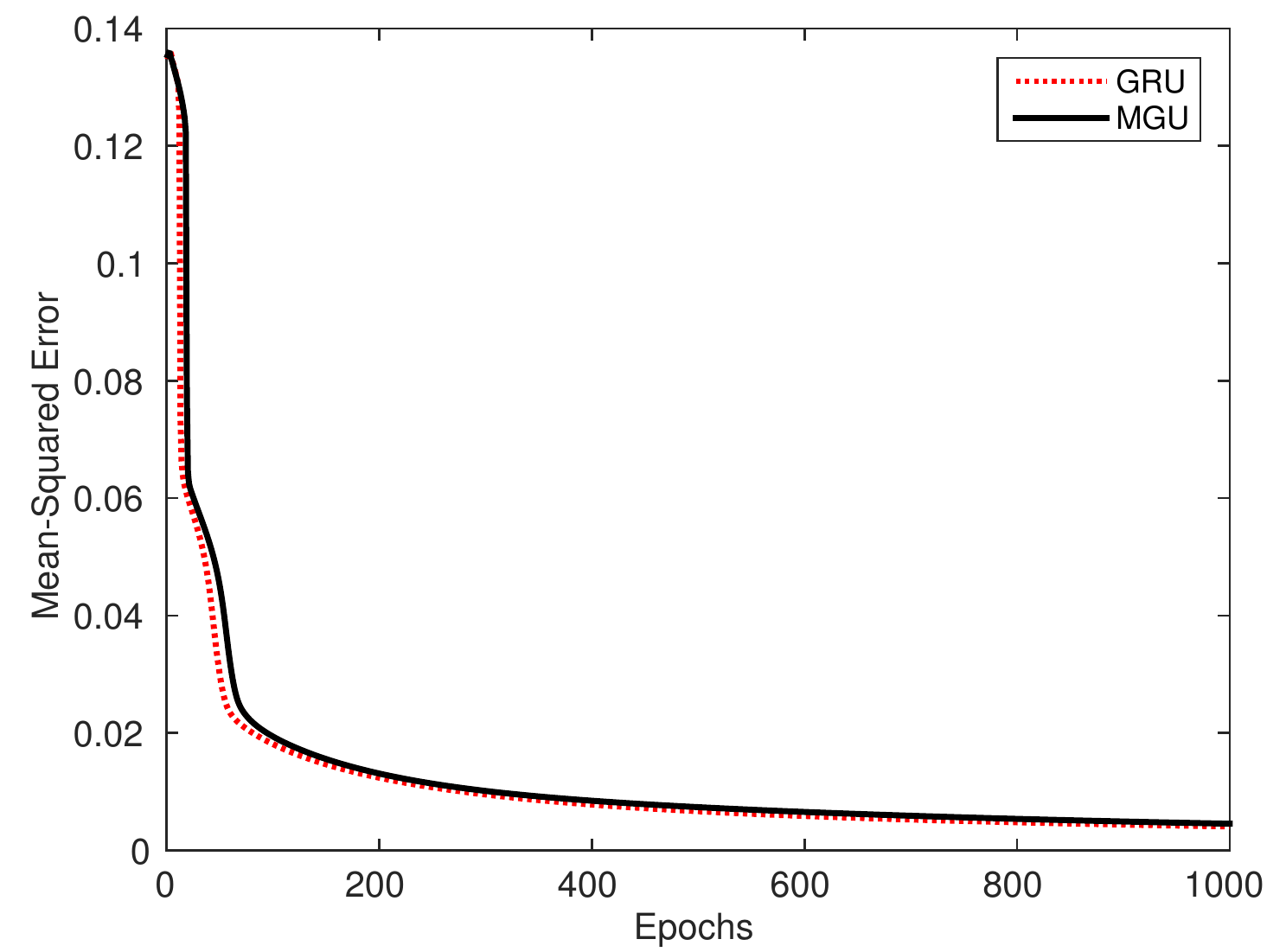}
 \caption{Test set mean squared error comparison (MGU vs. GRU) on the adding problem. Lower is better.} \label{fig:result_adding}
\end{figure}

We generated a training set of 10,000 examples and a test set of 1,000 examples. In Figure~\ref{fig:result_adding}, we show the mean squared error of this simple regression task on the test set. MGU is slightly worse than GRU in the first 100 epochs. However, after 100 epochs these two algorithms have almost indistinguishable results. After 1,000 epochs, the mean squared regression error of both methods are below 0.005: 0.0041 for GRU and 0.0045 for MGU. 

This simple experiment shows that MGU can smoothly deal with sequence input with a moderate length around 50. And, MGU has fewer parameters than GRU: 41,400 (MGU) vs. 62000 (GRU).

MGU also trains faster, which takes on average 6.85 seconds per epoch, while GRU requires 8.60 seconds.

\subsection{IMDB} \label{sec:results_imdb}

The second problem we study is sentiment classification in the IMDB movie reviews, whose task is to separate the reviews into positive and negative ones. This dataset was generated in~\cite{lr:Maas2011_IMDB}.\footnote{Available at \href{http://ai.stanford.edu/~amaas/data/sentiment/}{http://ai.stanford.edu/~amaas/data/sentiment/}.} There are 25,000 movie reviews in the training set, another 25,000 for testing. We use the provided bag-of-words format as our sequence input. The maximum sequence length is 128. Both MGU and GRU have 100 hidden units; batch size is 16, and the learning rate is $10^{-8}$ with a 0.99 momentum. Similar to the adding problem, we use a fully connected layer on top of the last hidden state to classify a movie review as either positive or negative.

\begin{figure}
 \centering
 \includegraphics[width=\columnwidth]{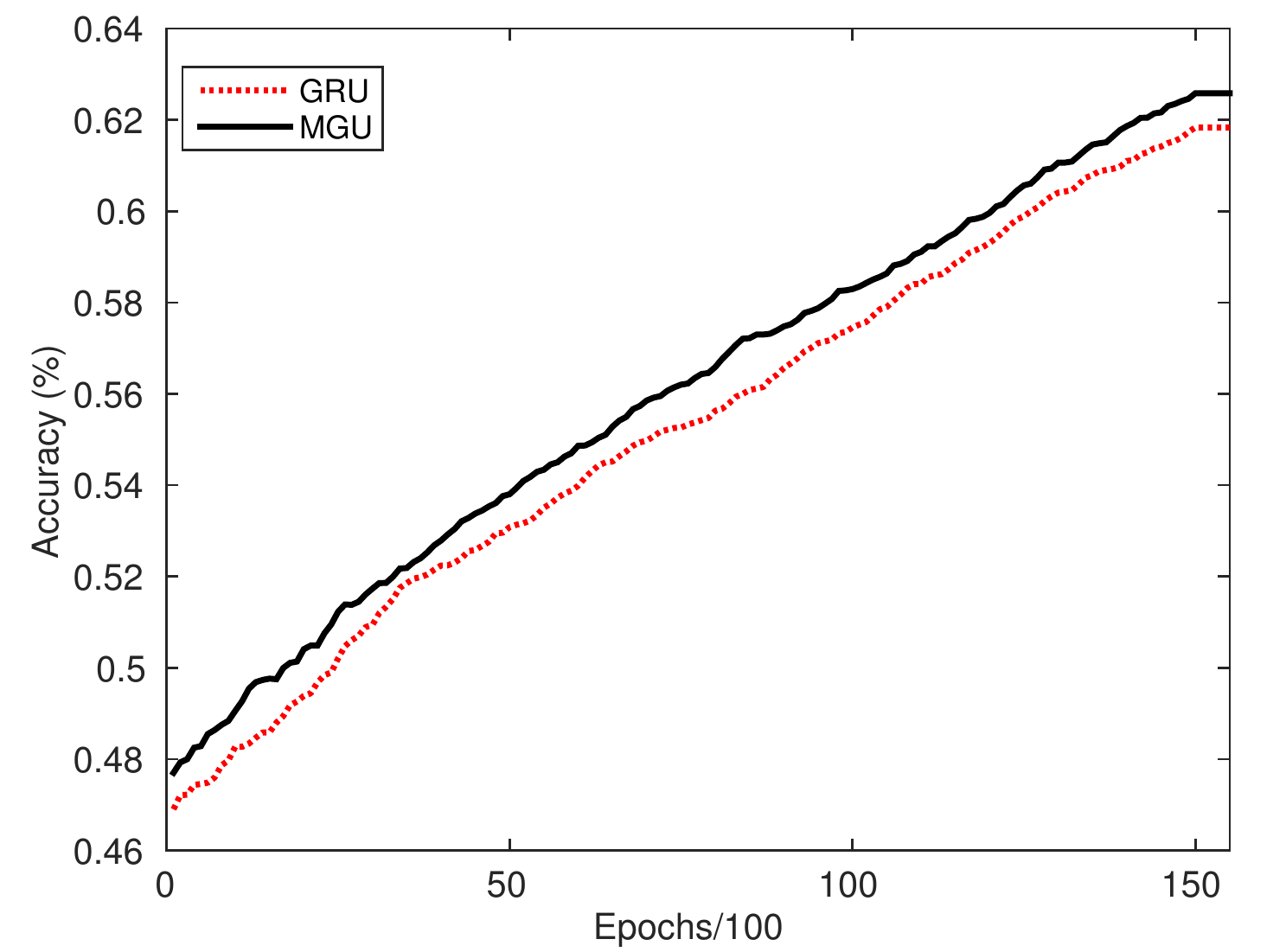}
 \caption{Test set classification accuracy comparison (MGU vs. GRU) on the IMDB dataset. Higher is better.} \label{fig:result_imdb}
\end{figure}

We show the accuracy of this binary classification problem in Figure~\ref{fig:result_imdb}, evaluated on the test set. In the $x$-axis of Figure~\ref{fig:result_imdb}, we show the epoch number divided by 100. Because both curves show that they converge after 15,000 epochs, it is too dense to show the results of the model after every training epoch.

MGU consistently outperforms GRU in this example, although by a small margin. After convergence, the accuracy of GRU is 61.8\%. MGU achieves an accuracy of 62.6\%, obtaining a 1.3\% relative improvement. The input of IMDB is longer than that of the adding problem. In this larger and longer dataset, MGU verifies again it has comparable (or slightly better in this case) accuracy with GRU.

Again, MGU has two thirds of the number of parameters in GRU. MGU has 20,400 parameters, while GRU has 30,600 parameters. However, MGU trains much faster than GRU in this problem. The average running time per epoch for MGU is 5.0 seconds, only 35\% of that of GRU (which takes 14.1 seconds per epoch on average).

\subsection{MNIST} \label{sec:mnist}

The MNIST dataset by~\citet{lr:LeCun1998} contains images of handwritten digits (`0'--`9''). All the images are of size $28 \times 28$.\footnote{Available at \href{http://yann.lecun.com/exdb/mnist/}{http://yann.lecun.com/exdb/mnist/}} There are 60,000 images in the training set, and 10,000 in the test set. The images are preprocessed such that the center of mass of these digits are at the central position of the $28 \times 28$ image. 

\begin{figure}
 \centering
 \includegraphics[width=\columnwidth]{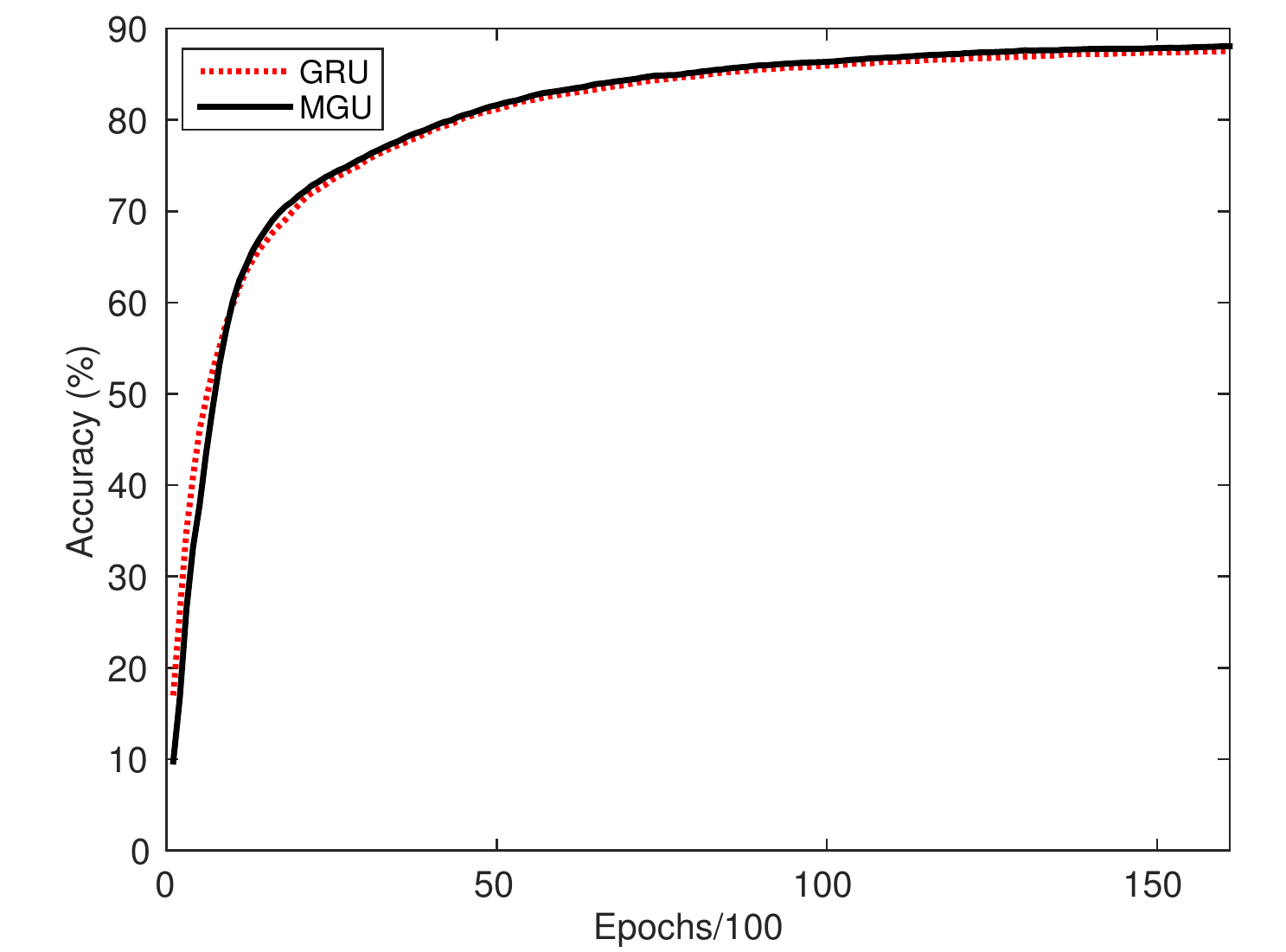}
 \caption{Test set classification accuracy comparison (MGU vs. GRU) on the MNIST dataset. This is the first task, where the sequence length is 28. Higher is better.} \label{fig:result_mnist}
\end{figure}

MNIST has been a very popular testbed for deep neural network classification algorithms, but is not widely used in evaluating RNN models yet. We use this dataset in two ways. The first is to treat each row (28 pixels) as a single input in the input sequence. Hence, an image is a sequence with length 28, corresponding to the 28 image rows (from top to bottom). For this task, we use 100 hidden units and the batch size is 100. The learning rate is $10^{-8}$ with a 0.99 momentum. A fully connected layer transfer the last row's hidden state into an output vector with 10 elements. Accuracy of MGU and GRU on the test set in this task are shown in Figure~\ref{fig:result_mnist}, where the $x$-axis is the epoch number divided by 100.

The performance on MNIST is similar to that on the adding problem (cf. Figure~\ref{fig:result_adding}), but the role of MGU and GRU has reversed. In the first 1,000 epochs ($x$-axis value up to 10), GRU is slightly better than our MGU. However, from then on till both algorithms' convergence, MGU's accuracy is higher than that of GRU. In the end (16,000 epochs), GRU achieves a 87.53\% accuracy, while MGU is slightly higher at 88.07\%. For this task, MGU has 25,800 parameters, while GRU has 38,700 parameters. The average per-epoch training time of MGU is 158 seconds, faster than that of GRU (182 seconds).

The second task on MNIST treats every pixel as one component in the input sequence. An image then becomes a sequence of length 784, with pixels scanned from left to right, and top to bottom. This task tests how MGU works when the sequence length is long. In this task, GRU has 30,600 parameters while MGU has 20,400. Other settings (learning rate etc.) are the same as those in the first task.

After 16,000 epochs, MGU's test set accuracy is 84.25\%, while with the same input length (784) and epoch number, IRNN's accuracy was below 65\%~\citep[see][Figure 3]{lr:Le2015_IRNN}. After 32,000 epochs, IRNN's accuracy was roughly 80\%, which is still worse than MGU's 84.25\% at 16,000 epochs. Note that when the number of epochs continued till 500,000, IRNN reached a high accuracy of above 90\%. Although we did not run MGU till this number of epochs, we have reasons to expect that MGU will similarly achieve a high accuracy if much longer training time is given. The accuracy of LSTM on this task is only around 65\% even after 900,000 epochs~\citep[see][Figure 3]{lr:Le2015_IRNN}.

The average training time per epoch of MGU is 48.1 seconds. However, GRU is much slower in this long sequence task, which takes 145.3 seconds per epoch.\footnote{We did not have enough time to wait for GRU to run to a high epoch number on this task. The accuracy of GRU is higher than that of MGU in the early training stage. However, MGU takes over after 710 epochs. This trend is similar to that of Figure~\ref{fig:result_mnist}.} If the same training time budget is allocated, MGU has significantly higher accuracy than GRU.

\subsection{Penn TreeBank} \label{sec:results_PTB}

\begin{figure}
 \centering
 \includegraphics[width=\columnwidth]{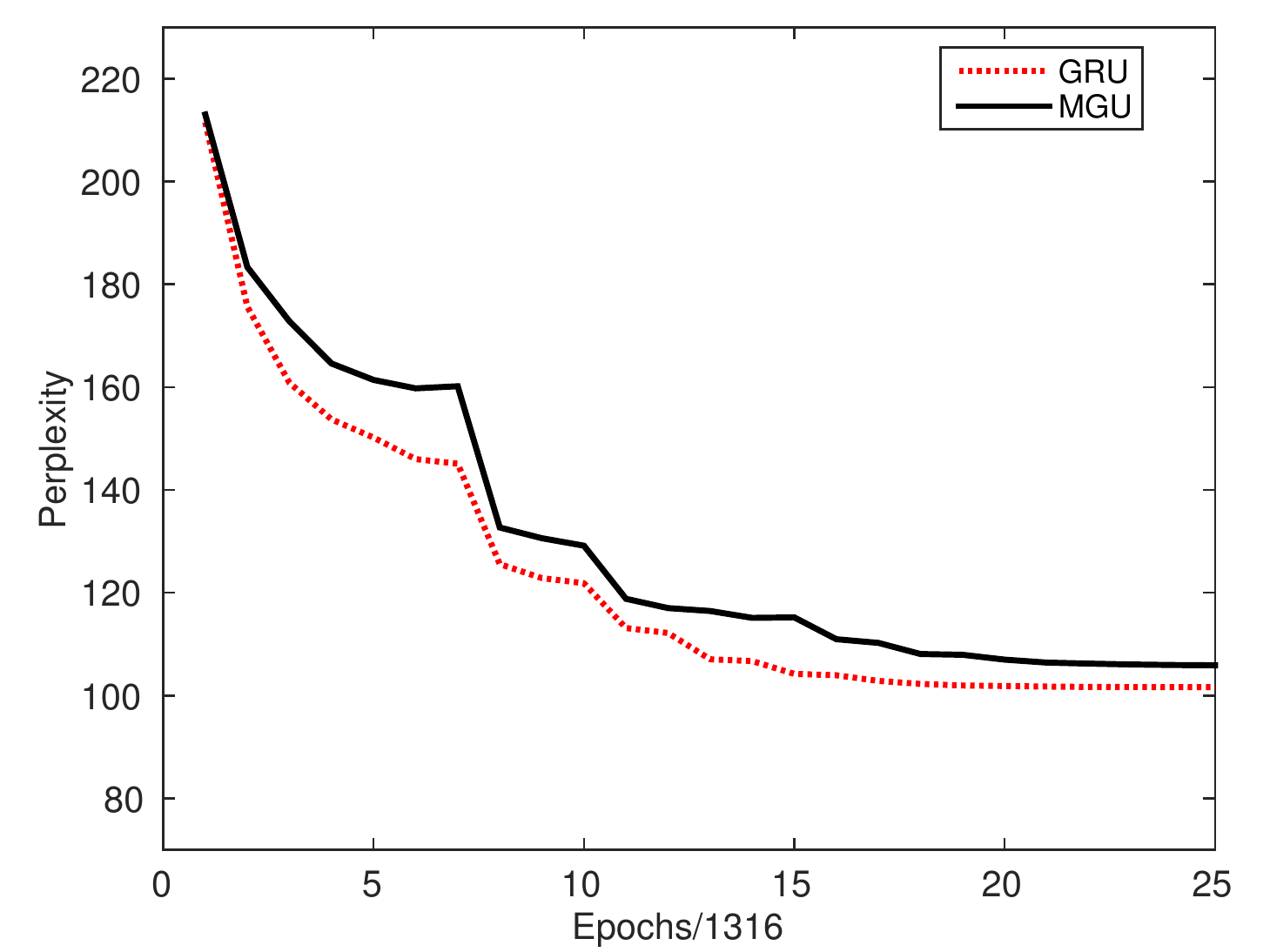}
 \caption{Test set perplexity comparison (MGU vs. GRU with 500 hidden units) on the PTB dataset. Lower is better.} \label{fig:result_ptb}
\end{figure}

The Penn TreeBank (PTB) dataset provides data for language modeling, which is released by~\citet{lr:Marcus1993_PTB}. For this dataset, we work on the word-level prediction task, which is the same as the version in~\cite{lr:Wojciech2014_RNN_regularization}.\footnote{Available at \href{https://github.com/wojzaremba/lstm/}{https://github.com/wojzaremba/lstm/}, or from \href{http://www.fit.vutbr.cz/˜imikolov/rnnlm/simple-examples.tgz}{http://www.fit.vutbr.cz/˜imikolov/rnnlm/simple-examples.tgz}.} There are 10,000 words in the vocabulary. It has 929K words in the training set, 73K in the validation set, and 82K in the test set. As in~\cite{lr:Wojciech2014_RNN_regularization}, we use two layers and the sequence length is 35. The batch size is 20, and the learning rate is 0.01. Because dropout is not used, we tested MGU and GRU on small networks, whose number of hidden nodes range in the set $\{50,100,200,300,400,500,600\}$. Without dropout, both units overfit when the number of hidden units exceed 500. A fully connected layer predicts one of the 10,000 words. As a direct comparison, we show the perplexity of MGU and GRU in Figure~\ref{fig:result_ptb} when there are 500 hidden units. The $x$-axis is the epoch number divided by 1,316.

GRU has a small advantage over MGU in this task. In the right end of Figure~\ref{fig:result_ptb}, GRU's perplexity on the test set is 101.64, while MGU's is higher at 105.59. We observe the same behavior on the validation set. We can also compare these two algorithms when they have roughly the same number of parameters, as was done in~\cite{lr:Chung2014_Eval_GRU}. The comparison results are shown in Table~\ref{tbl:comparison}.

\begin{table*}
 \centering
 \caption{Results and statistics of GRU and MGU on the Penn TreeBank dataset. The average per-epoch training time is in seconds. The best result in each column is shown in boldface.} \label{tbl:comparison}
 \begin{tabular}{|r|rr|rr|rr|rr|}
  \hline
  \#hidden units & \multicolumn{2}{c|}{\#parameters} & \multicolumn{2}{c|}{Per epoch time} & \multicolumn{2}{c|}{Validation perplexity} & \multicolumn{2}{c|}{Test perplexity} \\
  & GRU & MGU & GRU & MGU & GRU & MGU & GRU & MGU \\ \hline
  50  & 37,800 & 25,200 & 177 & 174 & 172.71 & 185.34 & 165.52 & 177.39 \\ 
  100 & 90,600 & 60,400 & 180 & 163 & 130.94 & 140.35 & 125.24 & 135.33 \\
  200 & 241,200 & 160,800 & 184 & 175 & 111.28 & 119.77 & 105.75 & 114.14 \\
  300 & 451,800 & 301,200 & 197 & 181 & 108.08 & 113.21 & 102.55 & 108.27 \\
  400 & 722,400 & 481,600 & 201 & 190 & \textbf{107.92} & 110.92 & 102.33 & 106.02 \\
  500 & 1,053,000 & 702,000 & 211 & 198 & 108.30 & \textbf{110.68} & \textbf{101.64} & \textbf{105.89} \\
  600 & 1,443,600 & 962,400 & 218 & 205 & 111.42 & 113.50 & 104.88 & 108.22 \\ 
  \hline
 \end{tabular}
\end{table*}

MGU has one third less parameters than GRU. Thus, the number of parameters are roughly the same for MGU with 500 hidden units and GRU with 400 hidden units. When we compare these two algorithms using these settings, the gap between GRU and MGU becomes smaller (102.33 vs. 105.89 on the test set, and 107.92 vs. 110.68 on the validation set). 

If we compare these two algorithms with the same amount of training time, MGU is faster than GRU. MGU with 500 units is roughly as fast as GRU with 300 units; and, MGU with 300 units is similar to GRU with 100 units. When the numbers of hidden units are same (e.g., 500), the proposed MGU can run more epochs than GRU given the same amount of training time, which we expect will continue to decrease the perplexity of MGU.

We can also compare MGU with the results in~\cite{lr:Jozefowicz2015_Eval_LSTM}. When there are 400 hidden units, the total number of parameters (including all layers) of MGU is 4.8M, which can be fairly compared with the ``5M-tst'' result in~\citep[Table 3]{lr:Jozefowicz2015_Eval_LSTM}, so is GRU with 300 units. Our GRU implementation (with 300 hidden units) has a test set perplexity of 102.55, lower than the GRU (with 5M parameters) result in~\cite{lr:Jozefowicz2015_Eval_LSTM}, which is 108.42 ($=\exp(4.684)$). The proposed MGU (with 400 hidden units) achieves a test set perplexity of 106.02, also lower than the GRU result in~\cite{lr:Jozefowicz2015_Eval_LSTM}.

The SCRN method has still fewer parameters than the proposed MGU. When there are 100 hidden units, MGU has 60,400 parameters. A similar SCRN architecture has 100 hidden units and 40 context units~\citep[see][Table 1]{lr:Mikolov2015_SCRN}, which has 48,200 parameters, amounting to roughly 80\% of that of MGU. On this dataset, however, SCRN seems to saturate at test set perplexity 115, because SCRN with 100 and 300 hidden units arrived at this same perplexity. MGU gets lower perplexity than SCRN on this dataset.

\subsection{Discussions}

We have evaluated the proposed MGU on four different sequence data. The comparison is mainly against GRU, while results of IRNN and SCRN are also cited when appropriate. The input sequence length ranges short (35, 50--55), moderate (128), and long (784). The sequence data range from artificial to real-world, and the task domains are also diverse.

The proposed method is on par with GRU in terms of accuracy (or error, or perplexity). Given its minimal design of one gate, MGU has only two thirds of the parameters of GRU, and hence trains faster in all datasets. However, in some problems (e.g., Penn TreeBank), GRU converges faster than MGU. Overall, through these experimental results we believe MGU has proven itself as an attractive alternative in building RNN.

\section{Conclusions and Future Work}

In this paper, we proposed a new hidden unit for recurrent neural networks. The proposed Minimal Gated Unit (MGU) has the minimal design in any gated hidden unit for RNN. It has only one gate (the forget gate) and does not involve the peephole connection. Hence, the number of parameters in MGU is only half of that in the Long Short-Term Memory (LSTM), or two thirds of that in the Gated Recurrent Unit (GRU). We compared MGU with GRU on several tasks that deal with sequence data in various domains. MGU has achieved comparable accuracy with GRU, and (thanks to the minimal design) trains faster than GRU.

Based on our evaluations, MGU could be readily used as the hidden unit in an RNN, which may reduce memory footprint and training time in some applications. More importantly, the minimal design will facilitate our theoretical analysis or empirical observation (e.g., through visualization) of RNN models, and enhance our understanding and facilitate progresses in this domain. A minimal design also means that there are fewer possibilities of producing variants (which will complicate the analysis).

Ample ways are possible to further this line of research. Beyond analysis and understanding, we will also run MGU with more epochs, in more diverse and complex tasks, and regularize MGU to improve its accuracy.

\bibliography{abbr,BibAll}
\bibliographystyle{icml2016}

\end{document}